\documentclass[10pt,twocolumn,letter]{article}

\usepackage{amsmath,amssymb,amsthm}
\usepackage{rotating}
\usepackage{balance}
\usepackage{bbm}
\usepackage{tikz}
\usepackage[left=2.54cm,right=2.54cm,top=2.54cm,bottom=2.54cm]{geometry}

\title{Space-filling Curves for High-performance Data Mining}
\author{Christian B\"{o}hm}
\date{Ludwig-Maximilians-Universit\"{a}t M\"{u}nchen, Munich, Germany, boehm@ifi.lmu.de}

\begin{document}
\newboolean{arxivdraft}
\setboolean{arxivdraft}{true}
\newboolean{stocdraft}
\setboolean{stocdraft}{false}
\maketitle

\begin{abstract}
Space-filling curves like the Hilbert-curve, Peano-curve and Z-order map natural or real numbers from a two or higher dimensional space to a one dimensional space preserving locality. They have numerous applications like search structures, computer graphics, numerical simulation, cryptographics and can be used to make various algorithms cache-oblivious. In this paper, we describe some details of the Hilbert-curve. We define the Hilbert-curve in terms of a finite automaton of Mealy-type which determines from the two-dimensional coordinate space the Hilbert order value and vice versa in a logarithmic number of steps. And we define a context-free grammar to generate the whole curve in a time which is linear in the number of generated coordinate/order value pairs, i.e. a constant time per coordinate pair or order value. We also review two different strategies which enable the generation of curves without the usual restriction to square-like grids where the side-length is a power of two. Finally, we elaborate on a few applications, namely matrix multiplication, Cholesky decomposition, the Floyd-Warshall algorithm, k-Means clustering, and the similarity join.
\end{abstract}
\section{Introduction}
\label{sec:intro} \noindent Countless algorithms from data analysis, basic math \cite{DBLP:journals/toms/DongarraCHD90}, graph theory, etc. are formulated as two or three nested loops which process a larger collection of objects. Let us for instance consider the simple algorithm of matrix multiplication $A=B\cdot C$ determining the entries $a_{i,j}$ of $A\in\mathbb R^{n\times m}$ by the rule:\vspace{-2mm}
\[a_{i,j} = \sum_k b_{i,k}\cdot c_{k,j}.\vspace{-2mm}\]
Since in C-like languages matrices are stored in a row-wise order, it is common practice to transpose $C$ before computing the scalar product $\sum_k b_{i,k}^{\phantom{\mathsf{\tiny{T}}}}\cdot c_{j,k}^{\mathsf{\tiny{T}}}$ to achieve a higher access locality:\vspace{-2mm}\\
\begin{minipage}{1\columnwidth}
\[\hspace*{2mm}\mbox{\textbf{for }} i:=0 \mbox{\textbf{ to }} n-1 \mbox{\textbf{ do }}\hspace{60mm}\vspace{-2mm}\]
\[\hspace*{6mm}\mbox{\textbf{for }} j:=0 \mbox{\textbf{ to }} m-1 \mbox{\textbf{ do}}\hspace{1.5mm}a_{i,j}^{\phantom{\mathsf{\tiny{T}}}}:=\sum_k b_{i,k}^{\phantom{\mathsf{\tiny{T}}}}\cdot c_{j,k}^{\mathsf{\tiny{T}}}\hspace{0.5mm};\hspace{60mm}\]
\end{minipage}
This algorithm essentially reads $B$ one time, row by row, from main memory into cache.
For each row of $B$ all the rows of $C^{\mathsf{\tiny{T}}}$ are read into cache, and combined with the current row $B_{i,*}$. Unless the complete matrix $C^{\mathsf{\tiny{T}}}$ fits into cache, this cyclic access pattern leads to a failure of the cache mechanism: with strategies like LRU (Least Recently Used), every row of $C^{\mathsf{\tiny{T}}}$ will be removed from cache before it can be re-used. As a consequence, we have a total of $n$ transfers of the complete matrix $C^{\mathsf{\tiny{T}}}$ from main memory to cache. We could make our algorithm \textbf{cache-conscious} \cite{DBLP:conf/sigir/AlabduljalilTY13} by an additional loop:\vspace{-2mm}\\
\begin{minipage}{1\columnwidth}
\[\hspace{2mm}\mbox{\textbf{for }} I:=0 \mbox{\textbf{ to }} n-1 \mbox{\textbf{ stepsize }} s\mbox{\textbf{ do }}\hspace{60mm}\vspace{-2mm}\]
\[\hspace{6mm}\mbox{\textbf{for }} j:=0 \mbox{\textbf{ to }} m-1 \mbox{\textbf{ do }}\hspace{60mm}\vspace{-2mm}\]
\[\hspace{10mm}\mbox{\textbf{for }} i:=I \mbox{\textbf{ to }} I+s-1 \mbox{\textbf{ do}}\hspace{1.5mm}a_{i,j}^{\phantom{\mathsf{\tiny{T}}}}:=\mbox{\Large $\sum$}_k \hspace{1.5mm}b_{i,k}^{\phantom{\mathsf{\tiny{T}}}}\cdot c_{j,k}^{\mathsf{\tiny{T}}}\hspace{0.5mm};\hspace{60mm}\vspace{1.5mm}\]
\end{minipage}
Provided that we have a single cache, large enough to store $s$ rows of $B$ and 1 row of $C^{\mathsf{\tiny{T}}}$, this strategy is dramatically better, 
because now we have to transfer $C^{\mathsf{\tiny{T}}}$ from main memory to cache only $\lceil n/s\rceil$ times while we still transfer matrix $B$ once.
\begin{figure*}[t]
  \centering
    \includegraphics[width=0.95\textwidth]{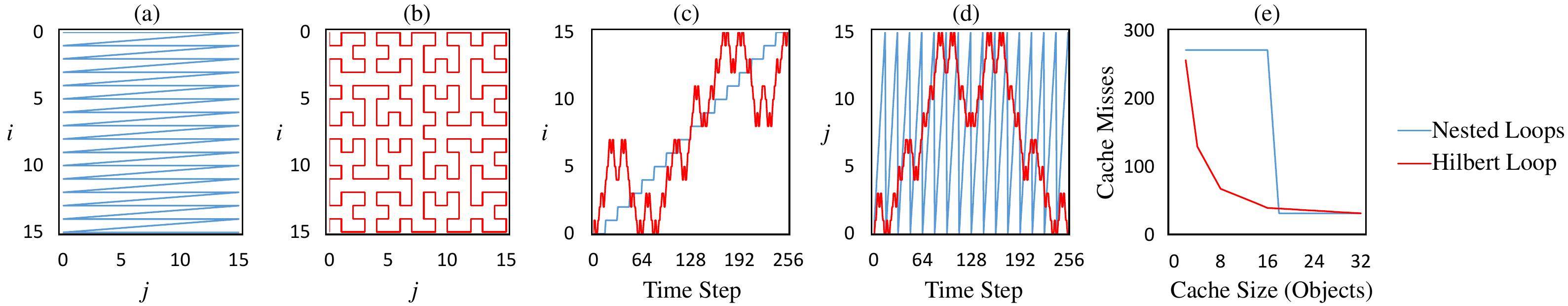}\vspace{-0mm}
   \caption{Comparison of the Traversal Order for Nested Loops (a) and Hilbert Loops (b). An improved locality can be recognized in the histories over time for variable $i$ (c) and $j$ (d), and a considerably improved cache miss rate (e).}\vspace{-4mm}
      \label{fig:motivation}
\end{figure*}
Modern processors support a memory hierarchy involving 2--3 levels of cache (L1, L2, L3, ordered by decreasing speed and increasing size), as well as a set of registers which are even faster than L1 cache. The main memory is usually organized as a virtual memory. Apart from expensive swapping to hard disk or solid state disk (if the matrices $B$ and $C^{\mathsf{\tiny{T}}}$ do not fit entirely into the physical main memory) 
we have to consider a second locality issue: the translation of virtual into physical addresses is supported by a very small associative cache called translation look-aside buffer. Only for a small number of pages this translation is fully efficient. While we might be able to determine the pure hardware size of all these cache mechanisms for a given hardware configuration it is difficult to know (and subject to frequent changes) how much of the various caches is available for our matrices, and not occupied e.g. by other concurrent processes or the operating system.

To efficiently support the complete hierarchy of memories of (effectively) unknown sizes, we need a different concept: a \textbf{cache-oblivious algorithm} \cite{DBLP:conf/focs/FrigoLPR99} is, unlike our above 3-loop construct, not optimized for a single, known cache size. It follows a strategy supporting a wide range of different cache sizes which can also be present simultaneously. The idea is to systematically interchange the increment of the variables $i$ and $j$ such that the locality of the accesses to both types of objects ($i$ \textbf{and} $j$) is guaranteed. Space-filling curves like the Hilbert curve or the Z-order curve act to some degree like the nesting of a high number ($2\cdot \log_2 n$) of loops going forward and backward with different step-sizes. 
In Figure~\ref{fig:motivation} we can recognize (a) the cyclic access pattern of nested loops, (b) the cache-oblivious access pattern of the Hilbert curve, (c) the histories of variable $i$ and (d) $j$ over time, and (e) the number of cache misses over varying cache size. We can see in Fig.~\ref{fig:motivation}(d) that the access pattern of the variable $j$ yields much more locality for the Hilbert loops compared to the cyclic access pattern of the nested loops. The result (e) is a dramatically improved number of cache misses, particularly for realistic cache sizes like 5-20\% of the main memory.

The main objective of this paper is to give an overview of our activities in the area of High-performance Data Mining, particularly about our variants of the Hilbert-curve and other space-filling curves to make such algorithms cache-efficient.

\section{Preliminaries on Space-filling Curves}\label{sec:preliminaries}
\begin{figure*}[t]
\ifthenelse{\boolean{arxivdraft}}{
\centering
\includegraphics[width=100mm]{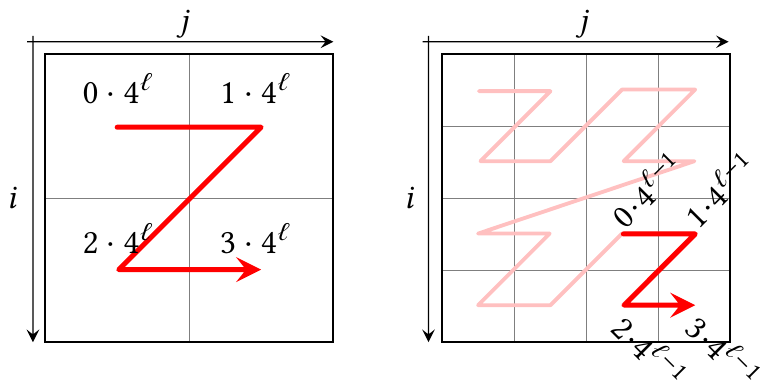}
}{
\hspace*{0mm}\includegraphics[width=73.8mm]{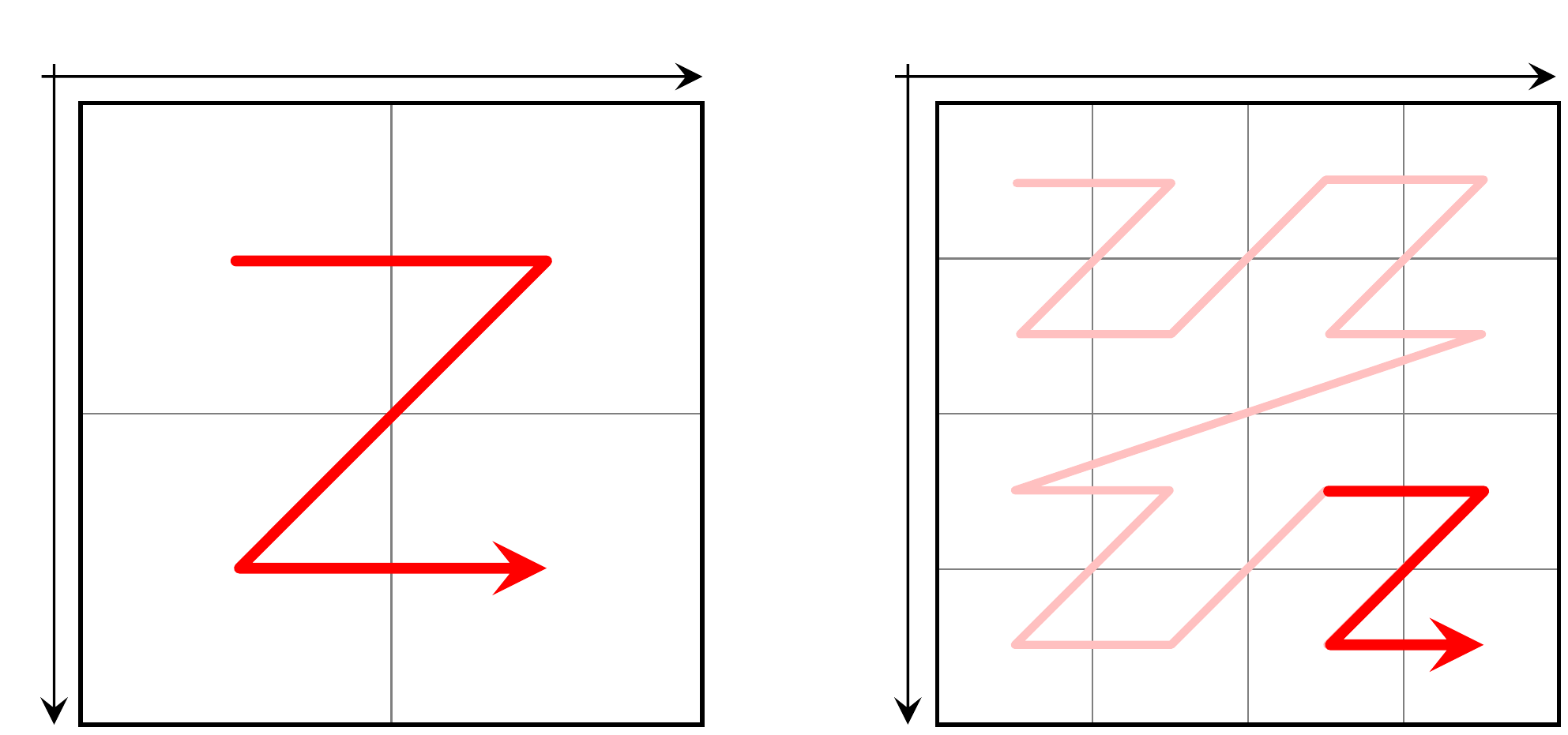}\\
\vspace{-41mm}
\[\setlength{\arraycolsep}{0mm}\begin{array}{@{\hspace{5.5mm}}l@{\hspace{6.5mm}}l@{\hspace{2.5mm}}l@{\hspace{3mm}}l@{\hspace{11.5mm}}l@{\hspace{16.5mm}}l@{\hspace{2.5mm}}l@{\hspace{-1.5mm}}l@{\hspace{20mm}}}
&&j&&&j&&\vspace{3mm}\\
&0\cdot4^\ell&&1\cdot4^\ell\vspace{6.5mm}\\
i&&&&i\vspace{-7.5mm}\\
&&&&&& \hspace*{-1mm}\mbox{\begin{turn}{45}$0\hspace{-0.6mm}\cdot\hspace{-0.6mm}4^{\ell-1}$\end{turn}}
& \mbox{\begin{turn}{45}$1\hspace{-0.6mm}\cdot\hspace{-0.6mm} 4^{\ell-1}$\end{turn}}\vspace{-2mm}\\
&2\cdot 4^\ell&& 3\cdot 4^\ell\vspace{3.6mm}\\
&&&&&& \hspace*{-0.5mm}\mbox{\begin{turn}{-45}$2\hspace{-0.6mm}\cdot\hspace{-0.7mm}4^{\ell-1}$\end{turn}}
& \hspace*{0.6mm}\mbox{\begin{turn}{-45}$3\hspace{-0.6mm}\cdot\hspace{-0.7mm} 4^{\ell-1}$\end{turn}}\end{array}\]
}
   \caption{Order Value Generation for Z-order $\mathcal Z(i,j)$.}
      \label{fig:zorder}
\end{figure*}

\noindent Classically, space-filling curves are defined as continuous, surjective mappings from the unit interval to the unit square in 2D or higher dimensional space \cite{cantor}. As our major focus are algorithms operating on object pairs which are numbered through indices $\in\{0, ..., n\}$ we define a space-filling curve $\mathcal C$ for this paper as a bijective mapping  $\mathcal C:\hspace{2mm}\mathbb N_0\times \mathbb N_0 \rightarrow \mathbb N_0$ assigning a \emph{pair of object indices} $(i,j)$ to an \emph{order value} $c$:
\[c:= \mathcal C(i,j); \hspace{8mm} (i,j):=\mathcal C^{-1}(c); \hspace{8mm} i,j,c \in \mathbb N_0.\]
Most space-filling curves like the Z-order, Hilbert curve, Gray-codes, etc. have been defined recursively in the 2D space of indices. This can be seen in Figure~\ref{fig:zorder}, left side, for the Z-order \cite{morton}: The space of indices is split into two halves in each dimension resulting in four partitions (quadrants) for 2D. The four quadrants are ordered (and numbered $\{0\cdot 4^\ell,...,3\cdot 4^\ell\}$) in a Z-shaped way. Note that the coordinate system is by convention oriented top down (and the second coordinate from left to right). Each partition is recursively split and ordered in the same way (Fig.~\ref{fig:zorder}, right side). The partitioning level $\ell$ represents the number of partitioning steps that still have to be done.

Straightforward recursive implementations of divide-and-conquer algorithms typically follow the Z-order implicitly. Usually space-filling curves are restricted to square-like spaces where the side length $n$ is a power of two ($i, j\in \{0,...,n-1\}$ where $n=2^\ell$) resulting in $n^2=4^\ell$ different order values $c\in \{0,...,4^\ell-1\}$, but we give an elegant solution to avoid this restriction.
\subsection{Z-order and Other Space-filling Curves}
\noindent Most space-filling curves agree with the Z-order in recursively bisecting the data space into $2\times 2$ or $3\times 3$ partitions, but try to increase the locality by avoiding the large jumps of the Z-order. In many cases, the patterns of the partitions resemble the shape of the main pattern but are not identical (e.g. rotated or reflected versions, like in the Hilbert curve). The Peano curve \cite{peano} partitions the space recursively in $3\times 3$ partitions with horizontally and/or vertically flipped sub-partitions.

We denote the functions computing the order values for a given pair $(i,j)$ of indices as follows:
\[\setlength{\arraycolsep}{0mm}\begin{array}{r@{\hspace{2.5mm}}l}
\mathcal C(i,j) & \mbox{for an arbitrary space-filling curve,}\\
\mathcal Z(i,j) & \mbox{for the Z-order, also called Morton-,}\\& \mbox{Lebesgue-, or N-order \cite{morton},}\\
\mathcal G(i,j) & \mbox{for Gray-codes \cite{roseman},}\\
\mathcal H(i,j) & \mbox{for the Hilbert curve \cite{Hilbert1935},}\\
\mathcal P(i,j) & \mbox{for the Peano curve \cite{peano},}\\
\mathcal N(i,j) & := \hspace{0.8mm}i\cdot n+j \hspace{2.8mm}\mbox{for the canonic order.}
\end{array}\]
We use the calligraphic letters for better visibility although $\mathcal C(i,j)$ denotes a function rather than a set. Further space-filling curves which are not elaborated in this paper include 
the Onion curve \cite{onion}. We also use the notation $\mathcal N(i,j):=i\cdot n+j$ for conventional nested loops in canonic order.

We denote by $(i,j):=\mathcal C^{-1}(c)$ the corresponding inverse function, and by $c:=\mathcal C^{\mathsf T}(i,j)=\mathcal C(j,i)$ its transpose (indices exchanged).

\subsection{Value Generation by Finite Automata}
\noindent To determine for a given pair $(i,j)$ of indices the corresponding Z-order value $\mathcal Z(i,j)$ we perform bit-interleaving: we consider the number $i$ as bit-string $i=\langle i_L ... i_2\,i_1\,i_0\rangle$ and likewise $j$, zero-filling from left (``zero-padding'') the shorter string to the length of the longer. The Z-order value is determined as the interleaved digits: $c=\langle i_L\,j_L ... i_1\,j_1\,i_0\,j_0\rangle$. Essentially, for the other space-filling curves this is done analogously (considering digits from a 3-adic system for the Peano curve), but the digits are translated e.g. by the use of a Mealy Automaton (i.e. a deterministic finite automaton producing a string of digits as output during the state transitions, \cite{mealy1955method}). The state transitions of the Mealy Automaton are labeled with a pattern 
$(i_\ell,j_\ell)\rightarrow c_\ell$ where $i_\ell$ and $j_\ell\in \{0,1\}$ ($\ell \in \{0,...,L\}$) represent a bit pair of the input (from the bit-strings representing $i$ and $j$, respectively) and $c_\ell\in\{0,...,3\}$ a four-adic digit (bit pair) of the generated output. For an example of the Mealy Automaton of the Hilbert-curve cf. Figure~\ref{fig:hilbert-dfa}. The Z-order is in this context a trivial Mealy Automaton with only one state and the self-transitions labeled $(0,0)\rightarrow 0, (0,1)\rightarrow 1, (1,0)\rightarrow 2, (1,1)\rightarrow 3.$ 
The Mealy Automaton for the inverse function $(i,j)=\mathcal C^{-1}(c)$ to determine the index pair for a given order value is analogous with input and output exchanged $c_\ell\rightarrow(i_\ell,j_\ell)$. Bit-interleaving as well as the Mealy Automata have a time complexity of $O(\log n)$ since the input is processed bit by bit; for Z-order and some other curves the computation is possible in $O(\log(\max(i,j)))$ time. On modern hardware, bit-interleaving is sometimes supported by the assembler-instructions ``PEXT'' and ``PDEP'' (parallel bit extract/deposit in the Bit Manipulation Instruction Set 2 of INTEL).

\section{Finite Automata for Hilbert}\label{sec:mealy}
\noindent We give here the methods to compute $h:=\mathcal H(i,j)$ and $(i,j):=\mathcal H^{-1}(h)$ directly using deterministic finite automata. Each bisection into four quadrants corresponds exactly to the processing of a bit pair of the binary representation of $i, j,$ and $h$, and the four basic patterns can be represented as states of a finite automaton which are analogously labeled as $U,D,A$ and $C$. These four letters are used because they represent the basic patterns to traverse the grid. The pattern $U$ starts in the upper left corner, goes one step down, one step to the left and one step right, ending in the upper right corner, like the shape of the letter ``U''. The pattern $D$ starts likewise in the upper left and traverses the grid like the round part of the letter ``D''. $A$ and $C$ start at the lower right corner drawing the letters reversely. The translation of the coordinate pair $(i,j)$ into the order value $h$ is computed bit by bit by taking a pair of binary digits $(i_\ell,j_\ell)\in\{0,1\}\times\{0,1\}$ as input and producing a digit $h_\ell\in\{0,1,2,3\}$ from a four-adic system (again equivalent to a bit pair) as an output during the state transition. A Deterministic Finite Automaton producing output during state transitions is called \emph{Mealy Automaton} or \emph{Mealy Machine} \cite{mealy1955method}.

\begin{figure}[t]
  \centering\vspace{0mm}\hspace*{1.2mm}
    \includegraphics[width=0.56\columnwidth]{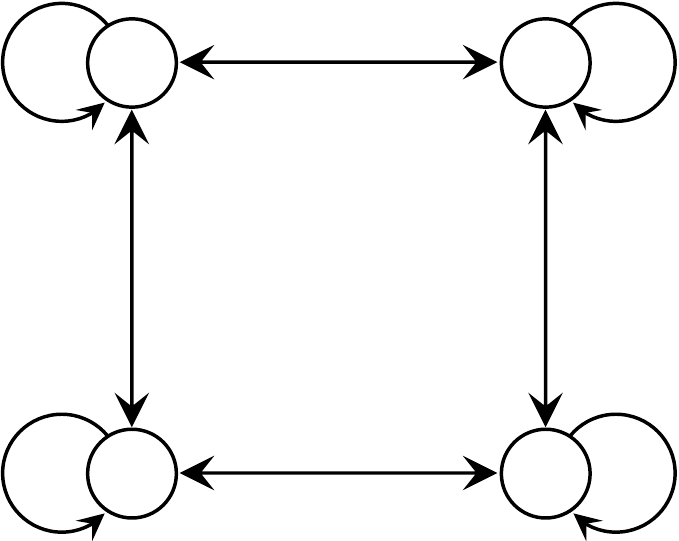}

    \vspace{-44.5mm}
    \[\begin{array}{c@{\hspace{2.2mm}}c@{\hspace{-0.6mm}}c@{\hspace{-0.6mm}}c@{\hspace{2.2mm}}c}
    \hspace*{-0.5mm}\begin{array}{c}(0,1)\rightarrow 1\vspace{-0.5mm}\\(0,0)\rightarrow 2\end{array}& A & \begin{array}{c}(1,0)\rightarrow 3\vspace{1.2mm}\\\phantom{(0,1)\rightarrow 1}\end{array} & D & \begin{array}{c}(0,1)\rightarrow 1\vspace{-0.5mm}\\(1,1)\rightarrow 2\end{array}\vspace{-0.5mm}\\
    &\mbox{ \begin{turn}{90}$\begin{array}{c}(1,1)\rightarrow 0\vspace{1.2mm}\\\phantom{(0,0)\rightarrow 0}\end{array}$\end{turn}}
    & & \mbox{ \begin{turn}{90}$\begin{array}{c}\phantom{(0,0)\rightarrow 0}\vspace{1.2mm}\\(0,0)\rightarrow 0\end{array}$\end{turn}} &\vspace{-1.4mm}\\
    \hspace*{-0.5mm}\begin{array}{c}(1,0)\rightarrow 1\vspace{-0.5mm}\\(0,0)\rightarrow 2\end{array}& C & \begin{array}{c}\phantom{(1,1)\rightarrow 1}\vspace{1.2mm}\\(0,1)\rightarrow 3\end{array} & U & \begin{array}{c}(1,0)\rightarrow 1\vspace{-0.5mm}\\(1,1)\rightarrow 2\end{array}\vspace{-0.5mm}
    \end{array}\]
   \caption{Hilbert Curve as a Mealy automaton.}
       \label{fig:hilbert-dfa}
\end{figure}

The state transition diagram is defined in Figure~\ref{fig:hilbert-dfa}. For each of the possible input bit pairs from $(i,j)$, it defines an output digit $\in\{0,1,2,3\}$ from a four-adic system, and a followup state. After bringing both bit strings to the same length by zero-padding the shorter one from left, we process these bit strings bit-pair by bit-pair, following the state transitions of the automaton.

We could easily select a basic resolution $L$ and consequently use one of the states as start point. In this case we would always consider $L$ digits of $i$ and $j$, which maybe include heading zeroes from the left. Our Mealy Automaton would always make exactly $L$ state transitions and could not translate coordinates $\ge 2^L$. However, we can even avoid the use of a basic resolution as a consequence of the labelling of the state transition between $U$ and $D$ (and vice versa) which is labelled $(0,0)\rightarrow 0$. This label means that every pair of heading zeroes from $(i_\ell,j_\ell)$ is translated into a heading $0$ in the output, just toggling between the states $U$ and $D$. We can safely ignore all these heading zero-pairs if we use the correct starting state out of $U$ or $D$ (decided by the parity of the length of the longer bit string). Alternatively, we can use the starting state $U$ \emph{always} and consider at most one additional heading zero-bit to make both lengths of the bit strings \emph{even}. The effective number of considered bits, $L(i,j)$ equals $\lceil\log_2(\max(i,j))/2\rceil\cdot 2$, with the guarantee to start with the same state as we would have done with every choice of $L\ge L(i,j)$.

The Mealy Automaton for the inverse $(i,j) := \mathcal H^{-1}(h)$ is completely analogous to that of $\mathcal H(i,j)$ but with input and output exchanged, e.g. $0\rightarrow (0,0)$ for the transitions between $U$ and $D$. The output bit-pair corresponds to one bit to be appended to the bit-string representing $i$ and one bit appended to $j$. Appending e.g. a one to $i$ corresponds to the mathematical operation $i:=2i+1$. We have to start with state $U$ and an even number of four-adig digits from $h$, i.e. $L(h) = \lceil\log_4(h)/2\rceil\cdot 2$.

The time complexity of the Mealy Automata is logarithmic, $O(\log \max(i,j))$ and $O(\log h)$, respectively.
\section{Context-free Grammars for Hilbert}
We can easily implement algorithms in a cache-oblivious way using the inverse Mealy Automaton of Section~\ref{sec:mealy}:
\[\setlength{\arraycolsep}{1mm}\begin {array}{l}\mbox{\textbf{for} $h:=0$ \textbf{to} $n^2-1$ \textbf{do}}\\
\hspace*{2em}(i,j):=\mathcal H^{-1}(h)\hspace{0.3mm};\\
\hspace*{2em}\mbox{\textit{actual loop body for} }(i,j);\end{array}\]
However, the $O(\log h)$ overhead of $\mathcal H^{-1}(h)$ is prohibitive for many applications including our running example of the backslash operator on a triangular matrix, which is also stated in a remark of \cite{DBLP:conf/focs/FrigoLPR99}. Therefore, we define a Lindenmayer system \cite{lindenmayer} which can be implemented even with a constant time and space complexity per loop iteration (in worst-case analysis).
A Lindenmayer System is a context-free grammar with non-terminal symbols $A,C,D$, and $U$, analogous to the patterns in the Mealy automaton (and exactly generating these patterns). The context-free grammar comprises four production rules:
\[
\begin{array}{c@{\hspace{4mm}::=\hspace{3.5mm}\pi\hspace{3mm}|\hspace{3mm}}ccccccc}
U & D & \downarrow  & U & \rightarrow & U & \uparrow    & C \\
D & U & \rightarrow & D & \downarrow  & D & \leftarrow  & A \\
A & C & \uparrow    & A & \leftarrow  & A & \downarrow  & D \\
C & A & \leftarrow  & C & \uparrow    & C & \rightarrow & U
\end{array}\]
The terminal symbols $\pi, \uparrow, \downarrow, \leftarrow, \rightarrow$ have the following meaning:
\[\begin{array}{l@{\hspace{5mm}}l@{\hspace{3mm}}l}
\pi & \mbox{process pair $(i,j)$;} & \mbox{(applied at level $\ell=-1$)}\\
\downarrow & \mbox{go down one step;} & (i:=i+1)\\
\uparrow & \mbox{go up one step;} & (i:=i-1)\\
\rightarrow & \mbox{go right one step;} & (j:= j+1)\\
\leftarrow & \mbox{go left one step;} & (j:=j-1)
\end{array}\]
The symbol $\pi$ stands for the actual loop body of the host algorithm to process the pair $(i,j)$.

The space-filling curve is generated by producing a word of the CFG starting from a level $L$ and using $U$ or $D$ as starting symbol (depending on the parity of $L$; $U$ if $L$ is even). The word generation is implemented by four mutually recursive methods $U(\ell), D(\ell), A(\ell), C(\ell)$ which perform the operations associated with the terminals on the right-hand side and a recursive call of e.g. $D(\ell-1)$ upon every non-terminal on the right-hand side. The symbol $\pi$ is generated exactly for the call $A(\ell=-1), C(\ell=-1),$ etc. and corresponds to the action to be performed on the current pair $(i,j)$ at the body of the loop. The method enumerates all pairs $(i,j)\in \{0,...,2^L-1\}\times\{0,...,2^L-1\}$ in the Hilbert Order:
\[\begin{array}{llc}\mbox{\textbf{function }} U(\ell)\\
\hspace*{1.5em}\mbox{\textbf{if} } \ell=-1 \mbox{ \textbf{then}}\\
\hspace*{3em}\mbox{\textit{process object pair $(i,j)$};}&\mbox{/\hspace{-0.5mm}/}&\pi\\
\hspace*{1.5em}\mbox{\textbf{else}}\\
\hspace*{3em}D(\ell-1);&\mbox{/\hspace{-0.5mm}/}&D\\
\hspace*{3em}i\hspace{0.5mm}:=i+1;&\mbox{/\hspace{-0.5mm}/}&\downarrow\\
\hspace*{3em}h:=h+1;\\
\hspace*{3em}U(\ell-1);&\mbox{/\hspace{-0.5mm}/}&U\\
\hspace*{3em}j\hspace{0.2mm}:=j+1;&\mbox{/\hspace{-0.5mm}/}&\rightarrow\\
\hspace*{3em}h:=h+1;\\
\hspace*{3em}U(\ell-1);&\mbox{/\hspace{-0.5mm}/}&U\\
\hspace*{3em}i\hspace{0.5mm}:=i-1;&\mbox{/\hspace{-0.5mm}/}&\uparrow\\
\hspace*{3em}h:=h+1;\\
\hspace*{3em}C(\ell-1);&\mbox{/\hspace{-0.5mm}/}&C\\
\end{array}\]
The recursion depth is obviously $L+1$, and since $n=2^L$, the space complexity is $O(\log n)$. The number of recursive calls is bounded above by $\tfrac{4}{3} n^2$ (as a geometric series) and thus the time complexity is $O(n^2)$. For now, we have the restriction that we can produce only $n\times n$-loops where $n$ is a power of two but we have proposed strategies how to generate cache-oblivious loops without this restriction (i.e. a loop enumerating $(i,j)\in\{0,...,n\}\times\{0,...,m\}$ for arbitrary $n,m\in \mathbb N_0$ in Hilbert Order) at constant overhead only.
\begin{figure*}[t]
    \vspace{-3mm}
  \centering
    \hspace*{-20mm}\includegraphics[width=0.7\textwidth]{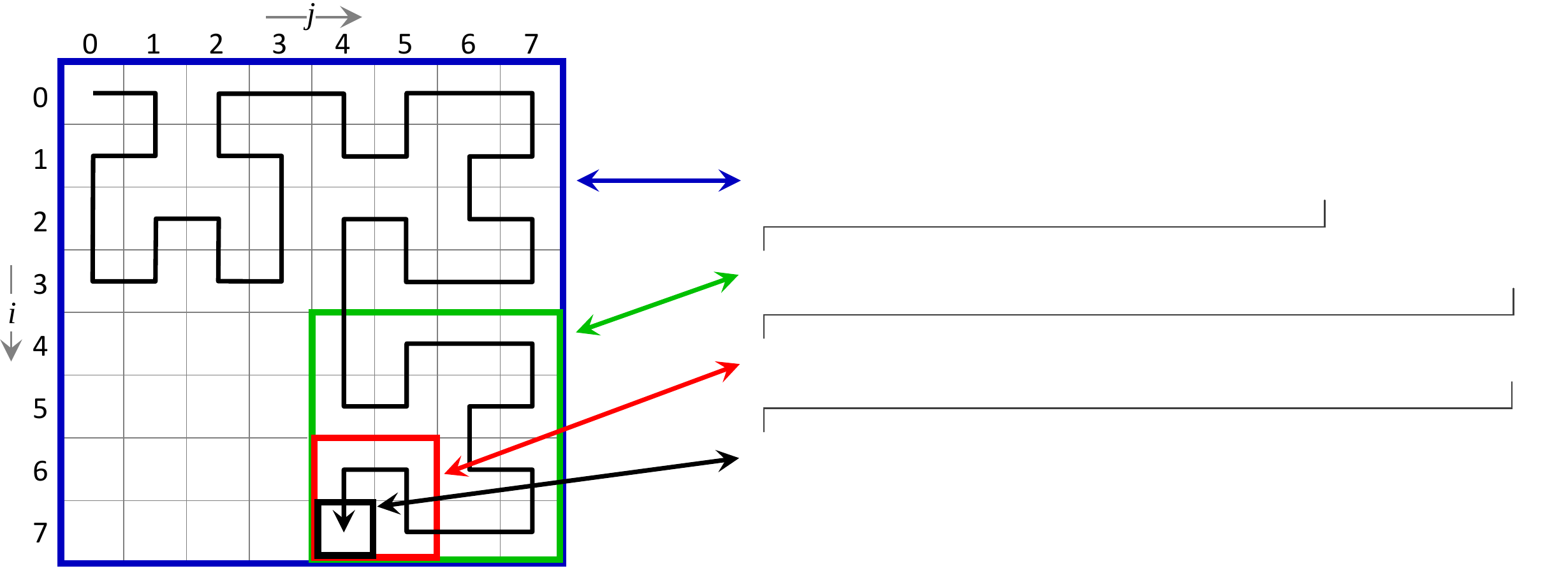}
    \vspace*{-43.3mm}\\
    \small{
\definecolor{green1}{rgb}{0,0.75,0}

\[\hspace*{50.5mm}\begin{array}{r@{\hspace{3.0mm}}r@{\hspace{3.0mm}}r@{\hspace{3.0mm}}r@{\hspace{9.5mm}}l}
\hspace{-3mm}^{\mbox{\scriptsize{labelX$_{0}$}}}\hspace{-1.7mm}\downarrow \hspace*{0.3mm} & \hspace{-3mm}^{\mbox{\scriptsize{labelX$_{1}$}}}\hspace{-1.7mm}\downarrow \hspace*{0.3mm} & \hspace{-3mm}^{\mbox{\scriptsize{labelX$_{2}$}}}\hspace{-1.7mm}\downarrow \hspace*{0.3mm} & \hspace{-3mm}^{\mbox{\scriptsize{labelX$_{3}$}}}\hspace{-1.7mm}\downarrow \hspace*{0.3mm} & \mbox{} \vspace{1mm}\\
\color{blue} D \hspace{4mm} \color{blue}::= \hspace{4mm} \color{blue}U & \hspace{0.5mm} \color{blue}\rightarrow \hspace{2.7mm} \color{blue}D & \hspace{0.5mm} \color{blue}\downarrow \hspace{2.7mm} \color{green1}D & \color{blue}\leftarrow \hspace{2.5mm}\color{blue} A & (\ell=2)\vspace{3.2mm}\\
\color{green1} D \hspace{4mm} \color{green1}::= \hspace{4mm}\color{green1} U & \hspace{0.5mm} \color{green1}\rightarrow \hspace{2.7mm} \color{green1}D & \hspace{0.5mm} \color{green1}\downarrow \hspace{2.7mm} D & \color{green1}\leftarrow \hspace{2.5mm} \color{red}A & \color{black}(\ell=1)\vspace{3.2mm}\\
\color{red} A \hspace{4mm} \color{red}::= \hspace{4mm} \color{red}C & \hspace{0.1mm} \color{red}\uparrow \hspace{3.2mm} \color{red}A & \hspace{0.5mm} \color{red}\leftarrow \hspace{2.7mm} \color{red}A & \color{red}\downarrow \hspace{2.5mm} \color{black}D & (\ell=0)\vspace{3.2mm}\\
\color{black} D \hspace{4mm} \color{black}::= \hspace{4mm}  \pi \hspace*{0.5mm}
\end{array}\]    }
    \vspace*{2mm}
   \caption{\vspace{1.5mm}Recursive Generation of Pairs $(i,j)$ Following the Hilbert-curve.}\vspace{-3mm}
      \label{fig:stack}
\end{figure*}

\section{Non-recursive Hilbert Value Generation}
\noindent In \cite{DBLP:conf/bigdataconf/BohmPP16,IEEE:transbigdata/BohmPP18}, we have proposed a method to enumerate the coordinate pairs and their Hilbert values in a non-recursive way. The basic idea is that all the information which is on the recursion stack of the mutually recursive functions $U(\ell), D(\ell), $ etc. can be recovered directly from the Hilbert value. While the derivation and proofs of the equivalence with the above CFG are lengthy, the result is astonishingly simple and shown in Figure~\ref{fig:nonrec}. The fundamental observation is that the level of the production rule which is responsible for a certain movement can be determined from the number of trailing zeros of the Hilbert values (after the increment). Modern hardware supports an assembler instruction to count heading and trailing zeros (\_tzcnt\_u64 counts the trailing zero bits of an unsigned 64-bit integer variable). If this is not available, we can also determine it with the binary logarithm:
\[\mbox{\_tzcnt\_u64}(h) = \log_2 (h\mbox{ \textbf{and}}_{\mbox{\scriptsize{bitw }}}{-h}).\]
From that, a variable $c\in\{0,1,2,3\}$ indicating the \underline{\textbf{c}}urrent direction of movement is updated with the following meaning:
\[\hspace*{4mm}\setlength\arraycolsep{0.6mm}\begin{array}{r@{\hspace{1.5mm}\Leftrightarrow\hspace{1.5mm}}lr@{\hspace{1.3mm}:=\hspace{1.3mm}}ccl}
c\hspace{-0.2mm}=\hspace{-0.2mm}0 & \mbox{look right:} & j & j & + &1,\vspace{-0.3mm}\\
c\hspace{-0.2mm}=\hspace{-0.2mm}1 & \mbox{look down:}  & i & i & + &1,\vspace{-0.3mm}\\
c\hspace{-0.2mm}=\hspace{-0.2mm}2 & \mbox{look left:}  & j & j & - &1,\vspace{-0.3mm}\\
c\hspace{-0.2mm}=\hspace{-0.2mm}3 & \mbox{look up:}    & i & i & - &1.\vspace{-1.5mm}
\end{array}\]
The coding of $c$ is chosen such that the increment of $i$ and $j$ can be implemented without pipeline-breaking ``if-else'' using the sign-preserving modulo operation: \[j := j+(c-1)\mbox{ \textbf{mod} }2; \mbox{ }i :=i+(c-2)\mbox{ \textbf{mod} }2.\]

Obviously, the algorithm performs inside its main loop only a constant number of operations, i.e. the overhead in each loop iteration is constant, in contrast to approaches which translate the Hilbert value into coordinates in each iteration. In addition, the algorithm uses only constant space, in contrast to the mutually recursive solutions. Moreover, the algorithm allows an elegant implementation of host algorithms, also facilitating compiler optimization: The whole algorithm of Figure~\ref{fig:nonrec} was implemented as a preprocessor macro which can be used like an ordinary loop instruction. 
\begin{figure}[b]
  \centering
  \vspace{-4mm}
  \fbox{
    \begin{minipage}{0.95\columnwidth}
{\scriptsize 1}\textbf{ function} LindenmayerNonRecursive()\\
{\scriptsize 2}      \hspace*{1.5em}$(i,j):=(0,0); h:=0; c:=3;$ \\
{\scriptsize 3}      \hspace*{1.5em}\textbf{while} $h < n^2$ \textbf{do}\\
{\scriptsize 4}      \hspace*{2.7em}\textit{process object pair $(i,j)$};\vspace*{-3mm}
      \[\hspace*{-3.2mm}\begin{array}{r@{\hspace{3em}}ll}
{\mbox{\scriptsize 5}}&        h\hspace{-2.5mm}    & :=h+1;\\
{\mbox{\scriptsize 6}}&        \ell \hspace{-2.5mm}& :=\lfloor\tfrac{1}{2}\cdot\mbox{\_tzcnt\_u64}(h)\rfloor + 1;\\
{\mbox{\scriptsize 7}}&        a \hspace{-2.5mm}   & :=\lfloor h/4^{\ell-1} \rfloor \mbox{ \textbf{mod} } 4;\\
{\mbox{\scriptsize 8}}&        c \hspace{-2.5mm}   & :=c\mbox{ \textbf{xor}}_{\mbox{\scriptsize{bitw}} } (3\cdot (\mbox{isOdd}(\ell \hspace{-0.5mm}-\hspace{-0.5mm} 1) \mbox{\textbf{xor} } a=3));\\
{\mbox{\scriptsize 9}}&        j\hspace{-2.5mm}    & :=j+(c-1)\mbox{ \textbf{mod} }2;\\
{\mbox{\scriptsize 10}}&        i\hspace{-2.5mm}    & :=\hspace{0.2mm}i\hspace{0.25mm}+(c-2)\mbox{ \textbf{mod} }2;\\
{\mbox{\scriptsize 11}}&        c \hspace{-2.5mm}   & :=c\mbox{ \textbf{xor}}_{\mbox{\scriptsize{bitw} } } (\mbox{isOdd}(\ell\hspace{-0.5mm} -\hspace{-0.5mm} 1) \mbox{ \textbf{xor} } a=1);\hspace{100mm}\vspace{0mm}
        \end{array}\]
    \end{minipage}
  }
  \caption{The Non-recursive Lindenmayer Algorithm.}
  \label{fig:nonrec}
\end{figure}
\section{Non-square Grids}
Usually space-filling curves are restricted to square-like spaces where the side length $n$ is a power of two ($i, j\in \{0,...,n-1\}$ where $n=2^\ell$) resulting in $n^2=4^\ell$ different order values $c\in \{0,...,4^\ell-1\}$. If we want to iterate over a non-square field $n\times m$ where $n\neq m$ or $n$ or $m$ do not agree with a power-of two, these space-filling curves yield a considerable overhead: the most obvious solution is to round-up to the next-higher power of two, i.e. determine \[N:=2^{\lceil \log_2(\max(n,m))\rceil},\]
iterate over an $N\times N$ grid using the space-filling curve, and ignore all pairs $(i,j)$ where $i\ge n$ or $j\ge m$. If $n\approx m$, we generate at most three times too many pairs but for $n>>m$ or $n<<m$ this overhead is unlimited high. In \cite{arbitrarySizeImage}, a solution explicitly putting a number of independent Hilbert-curves together has been proposed. Since these curves are not connected in a locality-preserving way, the cache-oblivious effect is lost at the connections. 

We have proposed two different solutions to this problem, (1) the overlay-grid, and (2) the jump-over operation, which are both proposed for the Hilbert curve but not limited to this space-filling curve. Basically these solutions can be combined with other  space-filling curves.

\subsection{Overlay-grids}
\noindent While the conventional Hilbert-curve recursively partitions the space in $2\times 2$ sub-partitions until elementary  $2 \times 2$-cells are reached (which is only possible when starting with an $n \times n$ grid where $n$ is a power of two), this solution allows at the lowermost level not only $2\times 2$ but also $2 \times 3, 2\times 4, 3\times 4,$ and $4\times 4$ elementary cells. In \cite{DBLP:conf/bigdataconf/BohmPP16} we have shown that this is always possible if $\tfrac{m}{2}<n<2m$, and cases of more severe asymmetry should be handled by placing independent curves side-by-side or above each other.

In \cite{IEEE:transbigdata/BohmPP18}, we have additionally shown that it is possible to maintain the fundamental property of the Hilbert curve to make only one step in $i$ or $j$-direction.

The advantage of this solution is that the constant overhead per loop iteration can still be guaranteed, but the grid to be iterated needs to be a square. More complex forms like triangles are not possible in this way. We have called our cache-oblivious loop the FUR-Hilbert-Loop (for \underline{\textbf{F}}ast and \underline{\textbf{U}}n\underline{\textbf{R}}estricted).

\subsection{Jump-over}
\noindent In \cite{DBLP:conf/sigmod/PerdacherPB19} we have proposed a solution that works even on more general forms, like triangles. The idea is not to ignore $(i,j)$-pairs out of the actual grid one-by-one but to decide for complete $2^\ell\times 2^\ell$ bisection quadrants of any level $\ell$ if they can be safely discarded. The search for a reentry-point of the grid may, however, need a logarithmic time complexity. In spite of this disadvantage, the jump-over solution is very general since it allows to iterate over more complex grids. In many applications, we need only $(i,j)$-pairs with $i<j$ (the lower left triangle of the square). We have particularly considered join operations where the actual part of the space is determined by more complex operations, depending on a hierarchical index structure. We have called our variant of the Hilbert-curve using jump-over FGF-Hilbert-Loop (for \underline{\textbf{F}}ast \underline{\textbf{G}}eneral \underline{\textbf{F}}orm).

In addition, it is an advantage of the jump-over solution that the $1:1$-relationsip between each order value and coordinate pair is maintained. The algorithm keeps track of the real Hilbert values while enumerating pairs. If we have some pairs with a special meaning, it might be necessary to identify them according to their order value. An example are graph algorithms which process node pairs. If a node-pair is connected by an edge, this node pair is processed in a different way than a non-edge. The decision whether $(i,j)$ is an actual edge as well as the management of the edges may be facilitated by determining the Hilbert values of the edges and maybe sorting the edges according to the Hilbert value.

\subsection{Nano-programs}

Nano-programs are tiny parts of the space-filling curves which are pre-computed and stored in an intelligent, compressed format to fit into 64-bit variables (and processor registers). For the overlay-solution, we stored all $\{0,1,2,3\}\times\{0,1,2,3\}$ sub-grids, each for all four different orientations. This approach allows us flexibly to generate the cells of the overlay grid. A second advantage which is also relevant for the jump-over solution is that nano-programs accelerate the speed of the curve generation, because reading out the movements from a variable is faster than performing the operations in Lines 6--11 of Figure~\ref{fig:nonrec}. It is also possible to define nano-programs for other space-filling curves like the Z-order or the Peano-curve, but the coding of movements must be adapted to the properties of the curve. Details can be found in \cite{DBLP:conf/bigdataconf/BohmPP16}.

\section{Applications}
\noindent In \cite{DBLP:conf/bigdataconf/BohmPP16,IEEE:transbigdata/BohmPP18} we have considered the following algorithms and made them cache-oblivious using the Hilbert-curve with the overlay grid (FUR-Hilbert-Loop):
\begin{itemize}
\item Matrix Multiplication,\vspace{-3mm}
\item k-Means Clustering,\vspace{-3mm}
\item Cholesky Decomposition,\vspace{-3mm}
\item Floyd-Warshall (transitive closure of a graph). 
\end{itemize}
For Cholesky Decomposition and transitive graph-closure, some data dependencies which are not compatible with the traversal of Hilbert have to be considered. The grid was decomposed into maximum parts which are compatible with an arbitrary traversal. Moreover, we used SIMD and MIMD parallelism (Single/Multiple Instruction Multiple Data), i.e. parallel threads on multiple cores and vectorization using instruction set extensions like AVX, AVX-2, and AVX-512 (Advanced Vector eXtension).

The FGF-Hilbert-Loop (using jump-over-operations) was used in \cite{DBLP:conf/sigmod/PerdacherPB19} for the Similarity Join. This database primitive combines pairs from a set of usually high-dimensional vectors based on a threshold of their (dis-) similarity (or distance), and is therefore a basic operation for many data mining algorithms. If the vectors are organized in a multidimensional index structure, only a certain part of all possible pairs qualifies as candidates for join results. We have demonstrated that the performance improves by considerable factors if these candidate pairs are accessed in Hilbert-order in a cache-oblivious way; for details cf.  \cite{DBLP:conf/sigmod/PerdacherPB19}. We used the FGF-Hilbert-Loop with jump-over operations that discarded parts of the data grid that were excluded according to the information in the directory of the index structure.

The K-Means algorithm was also considered in \cite{DBLP:conf/sdm/BohmPP17,DBLP:conf/icdm/BohmP15} but in a cache-conscious rather than a cache-oblivious way, again applying SIMD and MIMD parallelism. In \cite{DBLP:conf/icdm/PlantB10} we considered for the EM-Clustering algorithm (Expectation Maximization) a new concept for MIMD-parallelism and distributed algorithms called asynchronous model updates. Here, the frequency with which processes exchange their intermediate results (like centroids, covariance matrices) is optimized considering the traffic on network or bus connection. In \cite{DBLP:conf/bigdataconf/AltinigneliKRBP14,DBLP:conf/kdd/AltinigneliPB13,DBLP:journals/tlsdkcs/BohmNPWZ09,DBLP:conf/btw/BohmNPZ09} we considered various aspects of GPU processing (Graphics Processing Units) for applications like density-based clustering (DBSCAN) and SNP interactions (Single Nucleotide Polymorphism, mutations of the DNA).

\section{Conclusion}
\noindent In this paper we summarized our recent activities in the area of High-performance Data Mining with particular focus on cache-efficiency through space-filling curves. We emphasized on our methodology to overcome the most important drawbacks of the Hilbert-curve and many other curves, i.e. its logarithmic effort to compute coordinates from order values and their restriction to grid sizes which are powers of two. The code of our methods can be downloaded here:\\
{\hspace*{-0.0mm}\scriptsize https:\hspace{-0.3mm}/\hspace{-0.8mm}/\hspace{-0.3mm}dmm.dbs.if\kern0pt i.lmu.de\hspace{-0.3mm}/\hspace{-0.3mm}content\hspace{-0.3mm}/\hspace{-0.3mm}research\hspace{-0.3mm}/\hspace{-0.3mm}furhilbert\hspace{-0.3mm}/\hspace{-0.3mm}furhilbert.h}

\section*{Acknowledgment}

\noindent This work has been funded by the German Federal Ministry of Education
and Research (BMBF) under Grant No. 01IS18036A. The authors of this work take full responsibility for its content.

\bibliographystyle{plain}
\bibliography{bibliography}
\end{document}